# Diagnostic Method for Hydropower Plant Condition-based Maintenance combining Autoencoder with Clustering Algorithms

S. Jad[1, 2], X. Desforges[2], P.Y. Villard[1], C. Caussidéry[1], K. Medjaher[2]

[1] *Électricité de France Hydro Sud-Ouest, Toulouse, France*
(e-mail: samy.jad@edf.fr, pierre-yves.villard@edf.fr, christian.caussidery@edf.fr).

[2] *Laboratoire Génie de Production, Université de Toulouse,
Université de Technologie Tarbes Occitanie Pyrénées, Tarbes, France*
(e-mail: samy.jad@doctorant.uttop.fr, xavier.desforges@uttop.fr, kamal.medjaher@uttop.fr)

**Abstract**: The French company EDF uses supervisory control and data acquisition systems in conjunction with a data management platform to monitor hydropower plant, allowing engineers and technicians to analyse the time-series collected. Depending on the strategic importance of the monitored hydropower plant, the number of time-series collected can vary greatly making it difficult to generate valuable information from the extracted data. In an attempt to provide an answer to this particular problem, a condition detection and diagnosis method combining clustering algorithms and autoencoder neural networks for pattern recognition has been developed and is presented in this paper.

First, a dimension reduction algorithm is used to create a 2- or 3-dimensional projection that allows the users to identify unsuspected relationships between datapoints.

Then, a collection of clustering algorithms regroups the datapoints into clusters. For each identified cluster, an autoencoder neural network is trained on the corresponding dataset. The aim is to measure the reconstruction error between each autoencoder model and the measured values, thus creating a proximity index for each state discovered during the clustering stage.

## 1. INTRODUCTION

Maintaining the stability of a national electricity network is a complex activity involving stakeholders operating various tasks such as the production, the transportation or the distribution of electricity. Therefore, the maintenance of production assets plays a key role by ensuring their maximum availability during periods of power grid stress. By providing flexible, cheap and low-carbon electricity, HydroPower Plants (HPPs) are valuable assets for frequency modulation as intermittent renewable energy sources are integrated into the energy mix to meet the challenge of climate change (Farfan & Breyer, 2017). However, the ageing of Europe's HPPs may become a source of concern as in 2021, the average age of an HPP in Europe was 46 years (Quaranta et al., 2021). Given the low remaining hydropower potential capacity in Europe, 277 of the 350 GW known hydropower potential capacity in 2022 being already exploited (2022 Hydropower Status Report, 2022), modernizing Europe's HPPs may appear to be a prudent option.

For this reason, EDF has implemented a comprehensive plan to modernise its HPPs through the installation of supervisory control and data acquisition systems combined with a data management platform allowing the collection of years of sensor datapoints, enabling remote control and monitoring of its HPPs with 10 MW or more turbogenerators. To ensure greater flexibility and availability of HPPs, the research on maintenance paradigm in the industry is shifting from corrective and systematic maintenance to conditional and predictive maintenance (Kougias et al., 2019). The Prognostic and Health Management (PHM) framework, and in particular the data-driven approach towards health indicator prognostic such as remaining useful life may appear as an efficient way to enable conditional and predictive maintenance of complex systems (Atamuradov et al., 2017). In the particular case of HPPs, it may provide practitioners with helpful tools to detect faulty states of subsystems by reducing the large number of raw indicators to a few key ones.

For this reason, after reviewing the state of the art in fault monitoring and maintenance of HPPs, a diagnostic method based on the combination of autoencoder neural networks and clustering algorithms is proposed and its main results are presented and discussed.

## 2. STATE OF THE ART

The production process of electricity in HPPs involve the conversion of energy from hydraulic to mechanical through the hydraulic turbines, and then from mechanical to electrical through the generators and transformers for export according the electricity grid standards (Figure 1). Other key subsystems such as reservoirs, penstocks, surge tanks, programmable logic controllers, effectors of many types (valves, wicket gates, …) their control systems and more provide services to the HPP (Delliou, 2003).

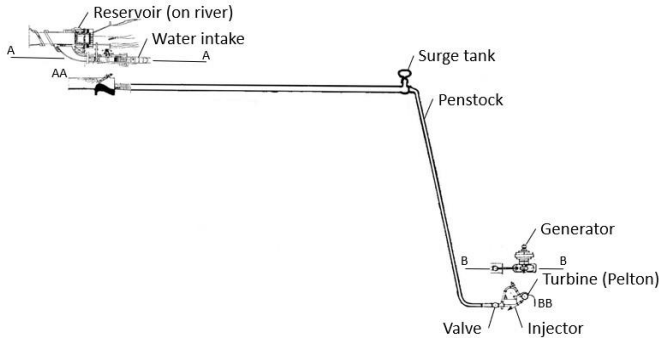

Figure 1. Schematic diagram of a HPP (run-of-the-river)

Therefore, a number of common failures in HPP's subsystems have been documented (Barbosa de Santis et al., 2021) and research has mainly focused on analytical or Finite Element studies of the concerned subsystems for water hammer (Bergant et al., 2006; Lupa et al., 2022), hydroacoustic vibration (Alligné et al., 2022; Héraud et al., 2019), erosion (Aumelas et al., 2016; Sangal et al., 2018), cavitation (Escaler et al., 2006), crack growth (Liu et al., 2016) and partial discharge (Sumereder, 2008) degradation modes leading to failures.

By contrast, recent works on predictive maintenance for HPPs have focused on the use of Machine Learning (ML) algorithms, which allow a more holistic approach by calculating health indicators that help to estimate degradation trends using Long-Short Term Memory (Hu et al., 2019; Velasquez & Flores, 2022), Self-Organising Map (SOM) (Betti et al., 2021) or Autoencoder neural networks (Hajimohammadali et al., 2023), with some being physics-informed by coupling ML algorithms with vibration analysis such variational mode decomposition (Fu et al., 2018; Wang et al., 2022), others providing information on HPP operation policies using SONARX neural network (Sacchi et al., 2004) or Support Vector Machine algorithm (Kumar & Saini, 2022).

Consequently, a good selection of sensors is necessary to enable monitoring of subsystems by practitioners, who may be able to detect early signs of known failure modes (Cheng et al., 2010). It is an essential part of the data acquisition tasks considered in the PHM framework together with data preprocessing. Data cleaning and feature extraction are performed in the latter and prepare the data for activities related to fault detection, diagnosis and prognosis. Decision-making and Human-Machine Interface rely on the results of diagnostic and prognostic tasks to suggest relevant actions to be taken (Atamuradov et al., 2017). As the first building blocks of the PHM framework have been achieved by EDF, it is possible to exploit sensor data through ML algorithms combined with knowledge of the system under consideration.

### 3. PROPOSED METHOD

For this reason, the datapoints manipulated in this paper are time series of averages values of analogue sensor signals recorded in company's data management platform at different frequencies, usually greater than one second to avoid saturation of data centres and bandwidth due to a single HPP. In fact, the number and types of sensors deployed varies greatly between systems to reflect the uniqueness of each HPP's subsystems, environmental conditions and failure types. For example, in the case studied, particular attention was paid to the penstock, effectors and generators, as shown by the types of analogue sensor signals available (Table 1), which main applications nowadays being early fault detection by comparison with tailored thresholds, retrospective fault analysis and production management.

Therefore, the proposal in this paper is a data-driven method for fault detection and diagnosis (Figure 2) by projecting the datapoints X into a lower dimensional representation in conjunction with clustering algorithms that aim at enabling practitioners to identify relationships between time series and real-life events that they would otherwise have struggled. In addition, the fitting of multiple autoencoder neural networks optimised on subsets related to the discovered clusters, seek to summarise the mass of collected information in a reduced number of key indices related to the proximity of healthy or faulty states.

Table 1. Analogue sensor signals exploited in the proposed method

| Sensor signals | Units | Subsystem |
|---|---|---|
| Active power | MW | Generator |
| Reactive power | MVA | Generator |
| Water flow | $m^3.s^{-1}$ | Penstock, Turbine |
| Injector opening | mm | Turbine |
| Rotation speed | RPM | Turbine, Generator |
| Temperature | °C | Turbine, Generator, heat exchanger |
| Vibration | mm/s | Turbine |
| Elevation | mNGF* or mm | Surge tank, Reservoir, Effector control chains |
| Pressure | Pa | Penstock, Effector control chains |

*mètre Nivellement Général de la France (General Ground Reference Metre of France) based on the French reference network IGN69

*3.1 Dimension reduction module*

First of all, the dimension of the input dataset is reduced to a 2- or 3- dimensional space in the Dimension Reduction (DR) module to generate a graphical representation of the datapoints in maps of clusters, using a DR algorithm that must be selected based on its performance to reproduce both linear and non-linear relationships between datapoints, to generate consistent representation with datasets composed of the same time series at new timestamps and to be optimised in relatively constrained time with an acceptable complexity both in the selection of hyperparameters and in the interpretation of the results for the practitioners.

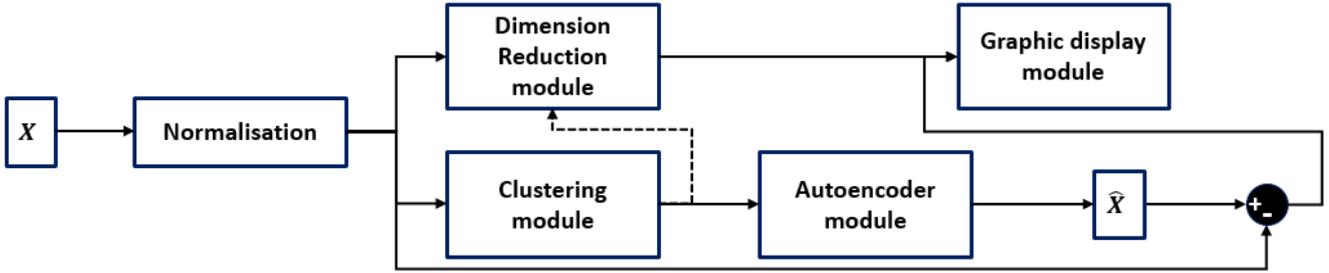

Figure 2. Proposed method for HPP condition monitoring

Since several types of DR algorithms proposed in the literature can meet the above criteria (Jia et al., 2022), they were tested during the experiments, the most notable ones being Principal Component Analysis (PCA), Multidimensional Scaling (MDS), Locally Linear Embedding (LLE), Isometric map (Isomap), T-distributed Stochastic Neighbours Embedding (T-SNE) and Uniform Manifold Approximation and Projection (UMAP).

Compared to the others, the latter fulfils the criteria for the datasets used in the experiments as it can reproduce non-linear relationships between datapoints as opposed to PCA and MDS, generate consistent representation when fed with never-before-seen datapoints from the same time series in contrast to T-SNE, and its optimization is faster LLE and Isomap.

### 3.2 Clustering module

In addition to the DR module, a clustering module consisting of different clustering algorithms applied to the same inputs is implemented to generate arrays of labels in order to assist the practitioner in identifying known faults based on timestamps, transient states during start and stop processes, power generation levels and possibly unknown faults. Datapoints are then highlighted in the DR representation and subsets are generated on which the autoencoder module is optimised in a second time according to each label.

However, as clustering algorithms are sensitive to different patterns in datasets, datapoints may be grouped into very different clusters (Bernard et al., 2021). The module is implemented by manually controlling the outputs of several clustering algorithms. During experiments, the results of SOM, K-Means, agglomerative clustering, Density-Based Spatial Clustering with Noises (DBSCAN) and its variant Hierarchical Density-Based Spatial Clustering with Noises (HDBSCAN) were implemented, the latter three requiring an additional layer based on a soft-voting classifier using K-Nearest Neighbour, Decision Tree and Support Vector Machine optimised on clustering results. The objectives are to benefit from non-parametrical algorithms as well as each algorithm's sensitivity to the topography of the datapoints.

### 3.3 Autoencoder module

In a second time, an auto-encoder is optimised on each subset identified using the clustering module outputs, therefore calculating a proximity score to clusters related to faulty states and deviation score from clusters related to healthy states (Betti et al., 2021), using a random search on a set of predetermined set of hyperparameters including number of layers and number of neurons on each layer, reproduced in mirror for the encoder and decoder parts, as well as activation functions common for all neurons in each parts (Figure 3).

As the input time series are normalised, the Mean Absolute Error (MAE) is computed as it is available for Python programming language and can be computed for series with numerous signals for regression-type problems similar to those encountered in the experiments.

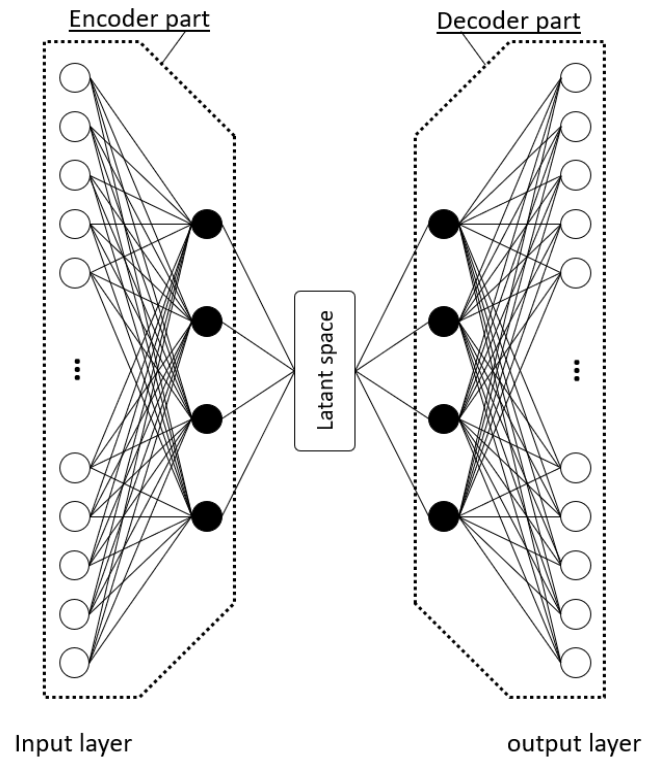

Figure 3. Example of autoencoder neural network with one hidden layer of four neurons in both encoder and decoder part

## 4. EXPERIMENTS, RESULTS AND DISCUSSION

Indeed, the time series used in the experiments were collected at a 10 minutes frequency between September 2018 and September 2023 at a run-of-the-river HPP in south-west France commissioned in 1954, which produces electricity from 2 turbogenerators of 10 and 20 MW respectively. These two turbogenerators share a single penstock, the 10 MW turbogenerator being designated as G1 by the company is equipped with 1 Pelton turbine while the second one, of 20 MW designated as G2, is equipped with 2 similar Pelton turbines. The implementation of sensors is roughly similar between the 2 turbogenerators except for the lack of vibration sensors on all axis on the hydraulic bearings of the smallest one. As a run-of-the-river HPP, it is able to produce electricity at different power rates for most of the year with an annual shutdown on July and August.

Consequently, the same preparation is applied to the four different applications that are formed, one for each turbogenerator, one with all available analogue sensors and one with analogue sensors not related to power generation, then excluding active power, water flow for each turbogenerator and injector position. First, minimal data preparation is performed prior to normalisation. This includes replacing missing and NaN data by padding, and filtering out anomalous data using a multilayer perceptron combined with a bandpass filter for injector position as failure of the position sensors during the annual outages appears as a gain. Then, normalisation is applied following equation (1) and dataset is divided into a training dataset with timestamps selected from September 2018 to September 2019 and a test data with the remainder.

$$Norm(X) = \frac{X - X_{min}}{X_{max} - X_{min}} \quad (1)$$

Therefore, the machine learning models in the DR, clustering and auto-encoder modules are optimised on the training datasets of the four applications through multiple runs to identify a satisfactory combination of hyperparameters. For UMAP models, these are mainly the number of epochs and computed neighbours, the initialisation algorithm, the minimum distance between datapoints in the resulting representation and the chosen metric for embedding matrix optimisation (McInnes et al., 2020). The aim is to produce a clear representation of the datapoints that allows clusters to be identified using timestamps and prior knowledge of the systems, time series gradients, with the help of the clustering module (Figure 4). As an example, agglomerative clustering and K-Means are able to group datapoints between operating and shutdown states with good precision of 99.93% for G1 and 99.41% for G2 data using K-Means, and 99.52% for G1 and 99,84% for G2 data using agglomerative clustering for a two-clusters objective, providing better reconstruction results by reducing the MAE of the autoencoder module (Table 2). In contrast, non-parametric clustering algorithms such as SOM, DBSCAN and HDBSCAN tend to generate a large number of clusters depending on the combination of hyperparameters (Bernard et al., 2021).

However, this step can be considered to be a tricky one as it requires human expertise and its inherent subjectivity to validate the clusters identified in output of both the DR and clustering modules. In addition, non-parametric clustering algorithms are not very good at recognising transients' clusters that occur under certain conditions such as weather complications, especially variations in outside temperatures and available water flow.

**Table 2. Evolution of MAE score for G1 and G2 datasets with and without clustering module**

| Dataset | MAE score |
|---|---|
| G1 K-Means HPP Shutdown | 0.03095 |
| G1 K-Means HPP Operating | 0.01889 |
| G1 Agglomerative Clustering HPP Shutdown | 0.02878 |
| G1 Agglomerative Clustering HPP Operating | 0.01901 |
| G1 no clustering | 0.04095 |
| G2 K-Means HPP Shutdown | 0.01551 |
| G2 K-Means HPP Operating | 0.02136 |
| G2 Agglomerative Clustering HPP Shutdown | 0.01609 |
| G2 Agglomerative Clustering HPP Operating | 0.02118 |

## 5. CONCLUSION

In the current situation, the preliminary results of the method appear to be encouraging for tasks such as identifying of operating and shutdown states and the estimating sensor signals by combining clustering algorithms with autoencoder neural networks to compute the reconstruction errors.

However, further development will be required to identify faulty states and more complex ones such as transient states of the HPP due to changes in production levels or large variations in external weather. The development of a method to automatically control and exploit the results of the different clustering algorithms, combined with the mapping of known transients states due to start and stop operations to optimise the autoencoder module on sequences of datapoints could provide more insight into the health state of the HPP and possibly remaining useful life calculation for predictive maintenance.

In addition, the development of a simpler and faster hyperparameter selection method can be considered as a necessary obligation for an industrial application.

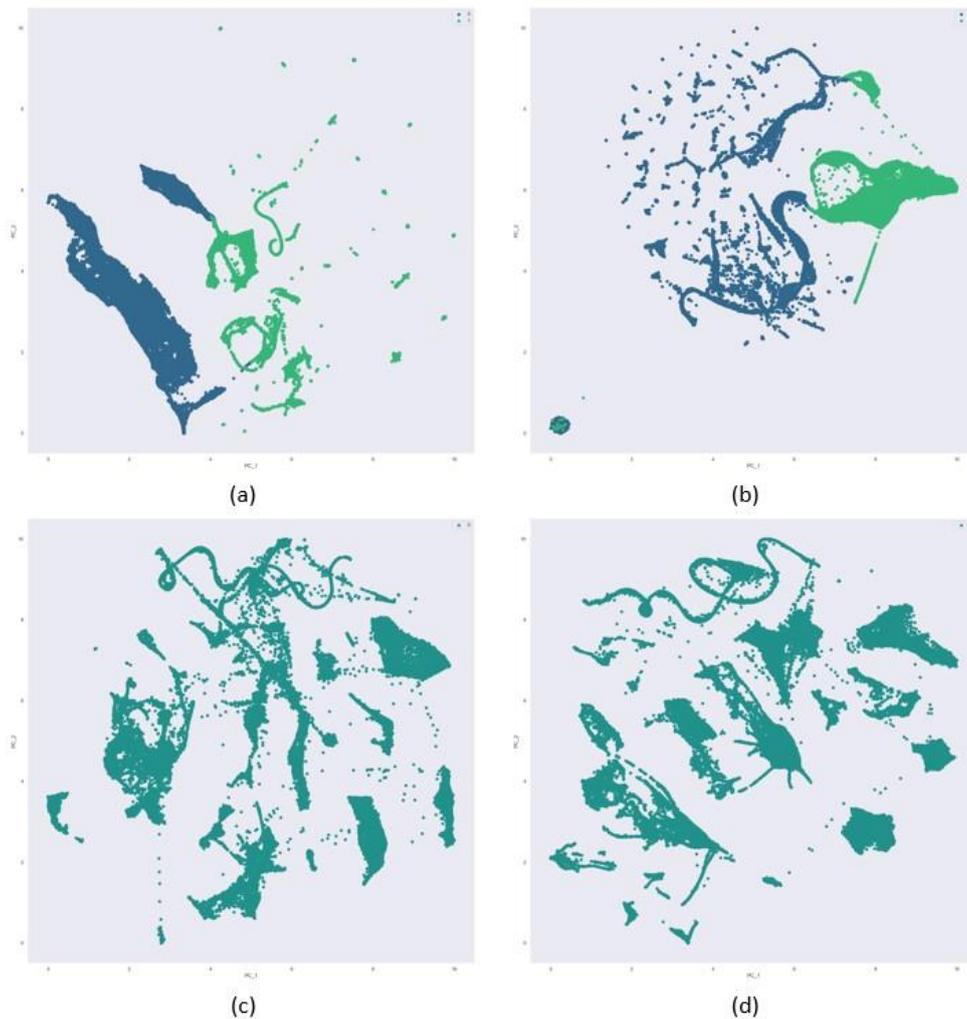

Figure 4. 2d UMAP representation: (a) G1 dataset (green: shutdown state, blue: operating state clustered with K-Means), (b) G2 dataset (blue: shutdown state, green: operating state with K-Means), (c) Dataset containing all data and (d) Dataset without active power data and related.